\documentclass[10pt,twocolumn,letterpaper]{article}

\usepackage{wacv}
\usepackage{times}
\usepackage{epsfig}
\usepackage{graphicx}
\usepackage{amsmath}
\usepackage{amssymb}

\usepackage[utf8]{inputenc}
\usepackage[percent]{overpic}
\usepackage[export]{adjustbox}
\usepackage[table]{xcolor}
\usepackage{multirow}
\usepackage{array}
\usepackage{enumitem}
\usepackage{caption}
\usepackage{dblfloatfix}

\makeatletter
\let\MYcaption\@makecaption
\makeatother

\usepackage[font=footnotesize]{subcaption}

\makeatletter
\let\@makecaption\MYcaption
\makeatother

\newcommand\Tstrut{\rule{0pt}{2.6ex}}         % = `top' strut
\newcommand\Bstrut{\rule[-0.8ex]{0pt}{0pt}}   % = `bottom' strut

\definecolor{limegreen}{rgb}{0.1961,0.8039,0.1961}
\definecolor{magenta}{rgb}{1,0,1}
\definecolor{cyan}{rgb}{0,1,1}

\def\wacvPaperID{541} % Enter the WACV Paper ID here

\wacvfinalcopy % *** Uncomment this line for the final submission

\ifwacvfinal
\def\assignedStartPage{1} % *** Enter the assigned starting page number (instead of 9876)
\fi

\usepackage[pagebackref=true,breaklinks=true,colorlinks,bookmarks=false]{hyperref}

\usepackage[sort&compress,capitalize,noabbrev,nameinlink]{cleveref}
\creflabelformat{equation}{#2#1#3}

\ifwacvfinal
\else
\fi

\begin{document}

%%%%%%%%% TITLE
\title{A Deep Temporal Fusion Framework for Scene Flow\\Using a Learnable Motion Model and Occlusions}

\author{René Schuster\textsuperscript{1} \hspace{0.5cm} Christian Unger\textsuperscript{2} \hspace{0.5cm} Didier Stricker\textsuperscript{1} \\
\textsuperscript{1}DFKI - German Research Center for Artificial Intelligence \hspace{5mm}
\textsuperscript{2}BMW Group \\
{\tt\small firstname.lastname@\string{dfki,bmw\string}.de}
}

%% omit the following lines up until the closing ``}''.
%% Additional authors and addresses can be added with ``\and'',
%% just like the second author.
%% To save space, use either the email address or home page, not both
%\and

\maketitle
%\thispagestyle{empty}

%%%%%%%%% ABSTRACT
\begin{abstract}
Motion estimation is one of the core challenges in computer vision. With traditional dual-frame approaches, occlusions and out-of-view motions are a limiting factor, especially in the context of environmental perception for vehicles due to the large (ego-) motion of objects. Our work proposes a novel data-driven approach for temporal fusion of scene flow estimates in a multi-frame setup to overcome the issue of occlusion. Contrary to most previous methods, we do not rely on a constant motion model, but instead learn a generic temporal relation of motion from data. In a second step, a neural network combines bi-directional scene flow estimates from a common reference frame, yielding a refined estimate and a natural byproduct of occlusion masks. This way, our approach provides a fast multi-frame extension for a variety of scene flow estimators, which outperforms the underlying dual-frame approaches.
\end{abstract}

%%%%%%%%% BODY TEXT
\section{Introduction} \label{sec:intro}

\begin{figure}[t]
	\setlength{\fboxrule}{1pt}
	\centering
	\begin{subfigure}[c]{0.9\linewidth}
		\includegraphics[width=\linewidth]{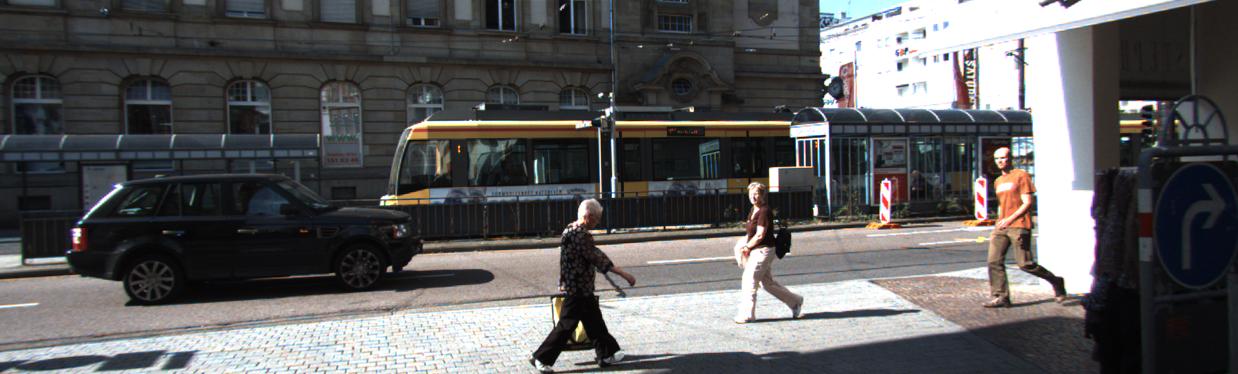}
		\caption{Reference Image}
		\vspace{1mm}
	\end{subfigure}
	\begin{subfigure}[c]{0.9\linewidth}
		\begin{overpic}[width=\linewidth]{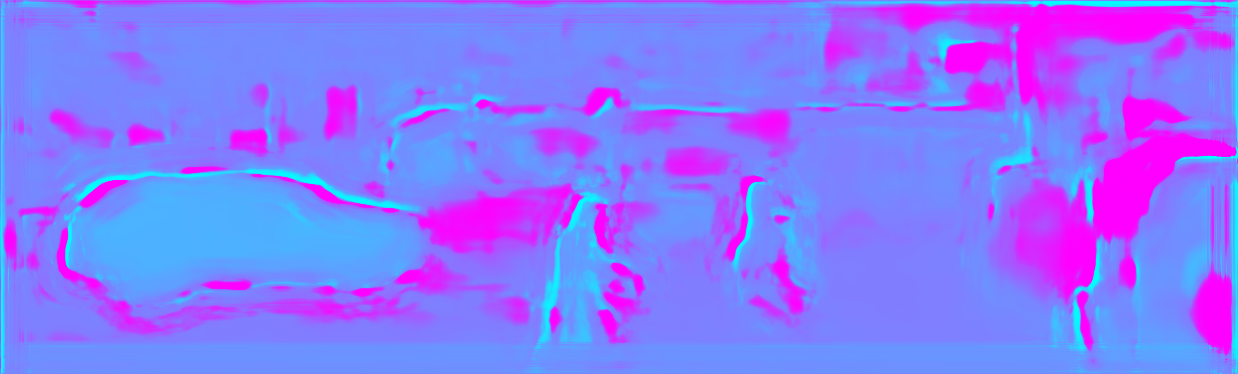}
			\put(1,25){\fcolorbox{black}{cyan}{\rule{3pt}{0pt}\rule{0pt}{3pt}} \small Forward}
			\put(1,20){\fcolorbox{black}{magenta}{\rule{3pt}{0pt}\rule{0pt}{3pt}} \small Backward}
		\end{overpic}
		\caption{Fusion Weights}
		\vspace{1mm}
	\end{subfigure}
	\begin{subfigure}[c]{0.9\linewidth}
		\begin{overpic}[width=\linewidth]{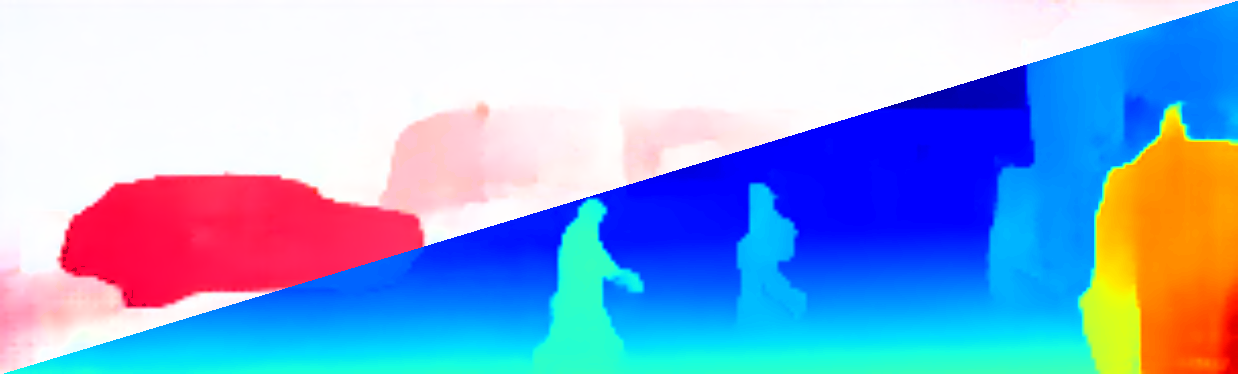}
			\put(1,25){\small SF outliers: 16.62 \% (occ: 66.46 \%)}
		\end{overpic}
		\caption{Dual Frame Result from \cite{saxena2019pwoc}}
		\vspace{1mm}
	\end{subfigure}
	\begin{subfigure}[c]{0.9\linewidth}
		\begin{overpic}[width=\linewidth]{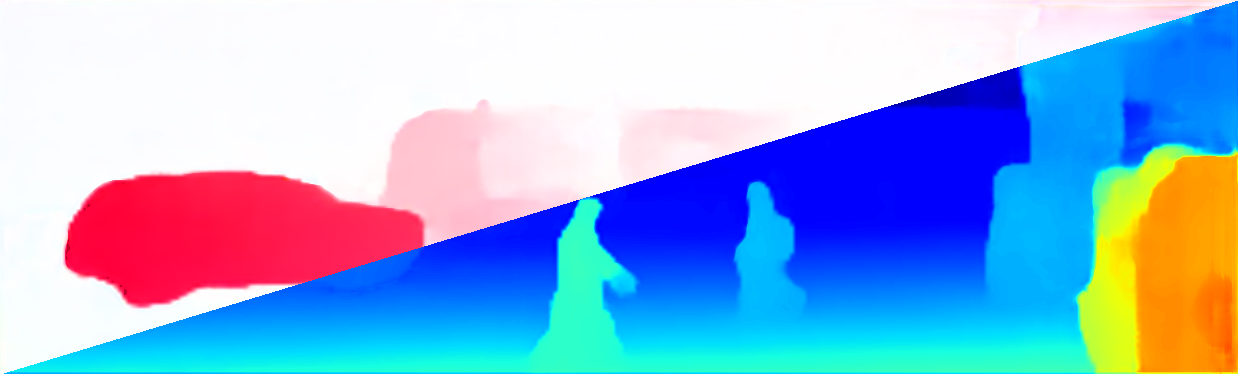}
			\put(1,25){\small SF outliers: 8.97 \% (occ: 8.75 \%)}
		\end{overpic}
		\caption{Our Fusion Result}
	\end{subfigure}
	\caption{Our deep temporal fusion (DTF) refines an initial dual-frame estimate by combination with an inverted backward scene flow. The fusion is realized as a pixel-wise weighted averaging and thus yields (soft) occlusion maps. This way, the initial results are significantly outperformed, especially in the difficult occluded areas.}
	\label{fig:teaser}
\end{figure}

The estimation of motion is important in many applications such as autonomous or assisted driving, robot navigation, and others.
A representation of motion in 2D image space (optical flow) is only a proxy for real world motion in the 3D world.
Scene flow is the estimation of 3D geometry and 3D motion and as such a much richer and realistic representation.
However, due to its higher complexity and its requirements on sensors, it is less often applied. 
Since the beginnings of scene flow estimation, major progress has been achieved.
Most recently, data-driven deep learning approaches have pushed the limits of scene flow estimation even further \cite{aleotti2020dwarf,jiang2019sense,ma2019drisf,saxena2019pwoc,yang2020upgrading}.
These approaches achieve state-of-the-art results at run times close to real time.
Yet, none of these deep learning methods utilizes a multi-frame setup which was shown to improve over a conceptually similar dual-frame approach for heuristic algorithms \cite{neoral2017object,schuster2020sffpp,taniai2017fsf,vogel2015PRSM}.
Many of these traditional, heuristic approaches use the additional information from multiple views as a kind of regularization during matching, making them more complex and reliable on specific, simplified motion models (\eg a constant motion assumption).
At the same time, all previous approaches (even multi-frame based) perform considerably worse in occluded areas (cf. \cref{tab:kitti}), which suggests that there is a lot of unused potential in multi-frame scene flow estimation.

More generic concepts for learning-based multi-frame settings were proposed in the context of optical flow \cite{liu2019selflow,maurer2018proflow,neoral2018continual,ren2019fusion}.
But these methods do not model the underlying issue of occlusions at all, or tackle the estimation of occlusions by bi-directional flow estimation (twice as much effort).

\begin{figure*}[t]
	\centering
	\includegraphics[width=1\linewidth]{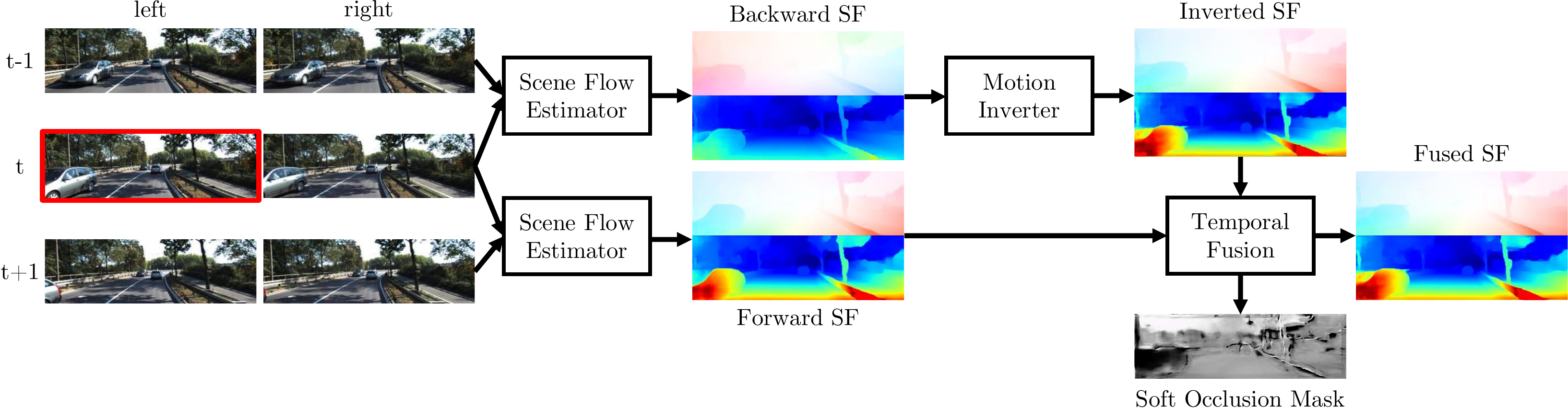}
	\caption{Overview of our proposed framework for deep temporal fusion with our trainable motion model.}
	\label{fig:framework}
\end{figure*}

In our work, we present the first deep fusion strategy for scene flow which is using a trainable, flexible motion model that exploits the geometric 3D information for self-supervised estimation of occlusion during temporal fusion (see \cref{fig:framework}).
Our framework overcomes some issues of previous work by the following contributions:
\begin{enumerate}[itemsep=2pt,topsep=4pt,leftmargin=*]
	\item It introduces a dedicated sub-network to temporally invert motion in the opposite direction of the target flow using a learned, flexible model of motion.
	\item It combines an initial estimate of forward scene flow with the inverted backward scene flow using a weighted average which results in the estimation of occlusions without explicit supervision.
	\item This way, the fused results show superior performance over the underlying dual-frame scene flow algorithms, especially in occluded areas.
\end{enumerate}
Additionally, our framework can be used together with any auxiliary scene flow estimator.

\section{Related Work} \label{sec:related}
\paragraph*{Scene Flow Estimation.}
The history of scene flow estimation began with early variational methods inspired by optical flow estimation \cite{huguet2007variational,vedula1999three}.
Many variants were presented for different sensor setups like RGBD \cite{herbst2013rgbd,jaimez2015primal}.
But all those methods are subjected to the requirements of the variational framework (small motions, good initialization) or of the hardware (\eg indoor environment for active depth cameras).
Within the classical sensor setup of stereo cameras, a big step forward was achieved by the introduction of the piece-wise rigid scene model \cite{behl2017bounding,lv2016CSF,menze2015object,vogel2013PRSF,vogel2015PRSM}.
However, these heuristic approaches presume local planarity and rigidity and lead to considerably long computation times.

A boost in run time was achieved with the introduction of the first deep learning algorithms due to the massive parallelization on GPUs.
At the same time, many of the newly proposed deep neural networks reached state-of-the-art results despite the lack of realistic, labeled training data \cite{aleotti2020dwarf,jiang2019sense,ma2019drisf,saxena2019pwoc,yang2020upgrading}.
Yet, no existing deep learning architecture for scene flow estimation makes use of the multi-frame nature of image sequences, which naturally exist in realistic applications.
Our approach fills this gap with a trainable, generic multi-frame solution for scene flow estimation.

Classical, heuristic approaches have shown that the transition from a single temporal frame pair to two (or more) is expected to improve the results \cite{neoral2017object,schuster2020sffpp,taniai2017fsf,vogel2015PRSM}.
However, all of these methods model the temporal relation of neighboring time frames as constant motion.
Our proposed framework distills a generic motion model from data.

\paragraph*{Deep Multi-Frame Models for Optical Flow.}
For optical flow there exists some previous work on deep multi-frame neural networks.
MFF \cite{ren2019fusion} computes forward flow for two consecutive time steps together with a backward flow for the central frame.
The backward flow is used to warp the previous forward motion towards the reference frame realizing a constant motion assumption.
A fusion network then combines the initial forward prediction and the warped one.
Occlusions are not modeled explicitly here.
ContinualFlow \cite{neoral2018continual} uses previous flow estimates as additional input during the estimation for the current time step.
Here, occlusions are learned as attention maps in a self-supervised manner similar to MaskFlownet \cite{zhao2020maskflownet} or PWOC-3D \cite{saxena2019pwoc}, but based on a cost volume instead of image features.
ProFlow \cite{maurer2018proflow} proposes an online inverter for motion that is trained for every frame on the fly.
In our work, we adopt this idea to avoid warping, but we only train a single inverter once to further avoid the re-training on every sample and the explicit estimation of occlusions at an early stage.
In SelFlow \cite{liu2019selflow} as in ProFlow also, occlusions are detected by a forward-backward consistency check.
SelFlow uses the additional multi-frame information by constructing cost volumes for forward and backward direction which are then used for the flow estimation.

Our work gets rid of any consistency checks, avoids warping to shift the handling of occlusions to a later stage, and learns a dedicated universal model for the inversion of motion.
Contrary to all mentioned cases, we propose a deep multi-frame model for the more complex problem of scene flow estimation.

\section{Deep Multi-Frame Scene Flow} \label{sec:method}
Consider a stream of stereo image pairs $I_t^l$ and $I_t^r$ for left and right camera at a given time $t$.
For our framework, we tackle the problem of scene flow estimation with respect to a reference view (left at time $t$) into the future (time $t+1$).
While dual-frame solutions only consider the four images at these two time steps, a multi-frame method incorporates information from at least one additional time (usually $t-1$ to avoid delay in the prediction and account for the symmetry in motion).
Our framework builds on this exact setup using three stereo pairs at time $t-1$, $t$, and $t+1$.
The idea is outlined in \cref{fig:framework} and can be summarized as follows.
We use an arbitrary auxiliary model for scene flow estimation to predict forward ($t \rightarrow t+1$) and backward ($t \rightarrow t-1$) scene flow with respect to our reference view. This avoids any form of warping and thus postpones the problem of occlusions.
Then, we learn a motion model that transforms the backward estimate into a forward motion.
Finally, a temporal fusion module combines the forward and transformed backward estimate to obtain a refined result.
For the fusion, we use a strategy of weighted averages.
This implicitly yields soft occlusion maps for the two motion directions without explicit supervision on occlusions.
The underlying dual-frame model that we mainly use is PWOC-3D \cite{saxena2019pwoc} due to its simple training schedule compared to other approaches.
However, in our experiments (\cref{sec:experiments:dual}) we show that our framework is not limited to this model.
The novel sub-networks for motion inversion and fusion are presented in more detail in the next sections.

\subsection{Temporal Scene Flow Inversion} \label{sec:method:inverter} 
\begin{figure}
	\centering
	\begin{subfigure}[c]{0.45\linewidth}
		\includegraphics[width=\linewidth]{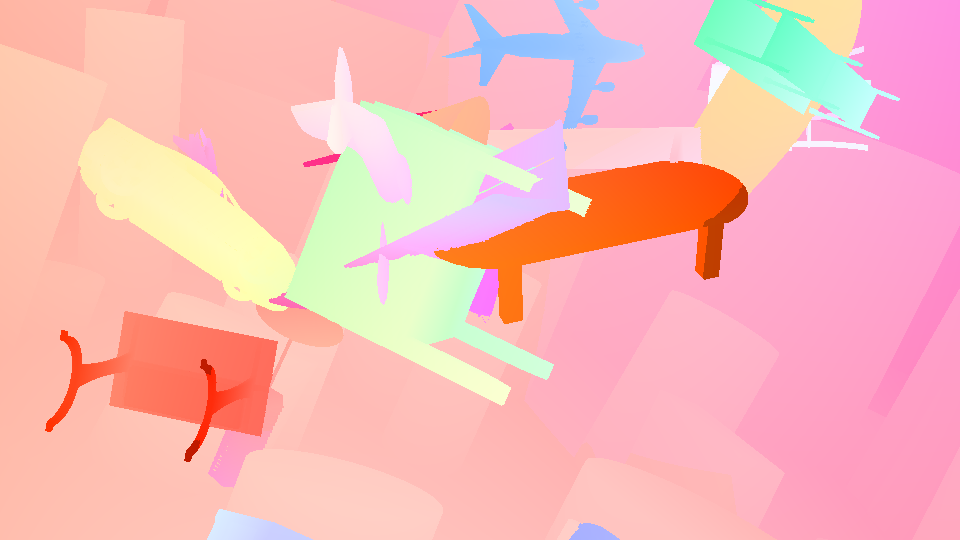}
		\caption{Backward Optical Flow}
		\vspace{1.5mm}
	\end{subfigure}
	\begin{subfigure}[c]{0.45\linewidth}
		\includegraphics[width=\linewidth]{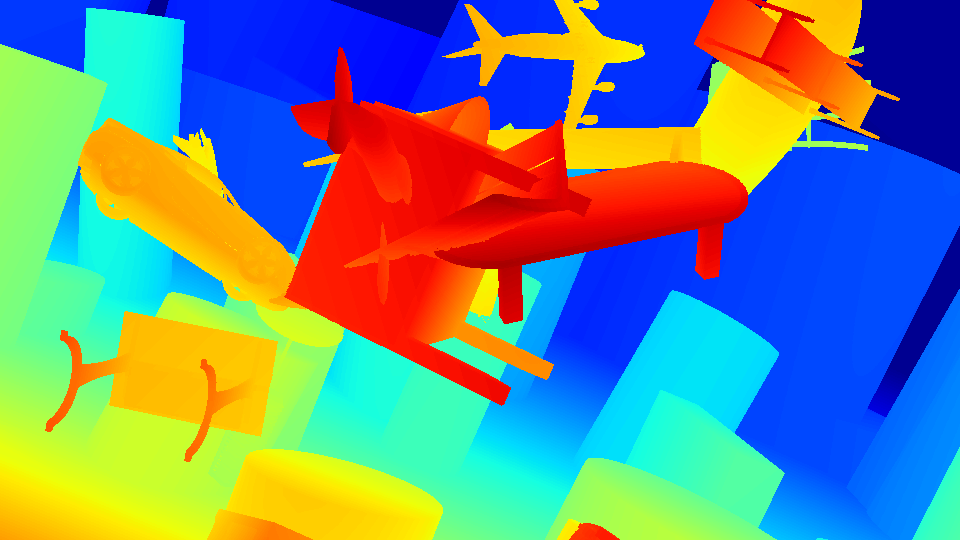}
		\caption{Disparity at $t-1$}
		\vspace{1.5mm}
	\end{subfigure}\\%
	\begin{subfigure}[c]{0.45\linewidth}
		\includegraphics[width=\linewidth]{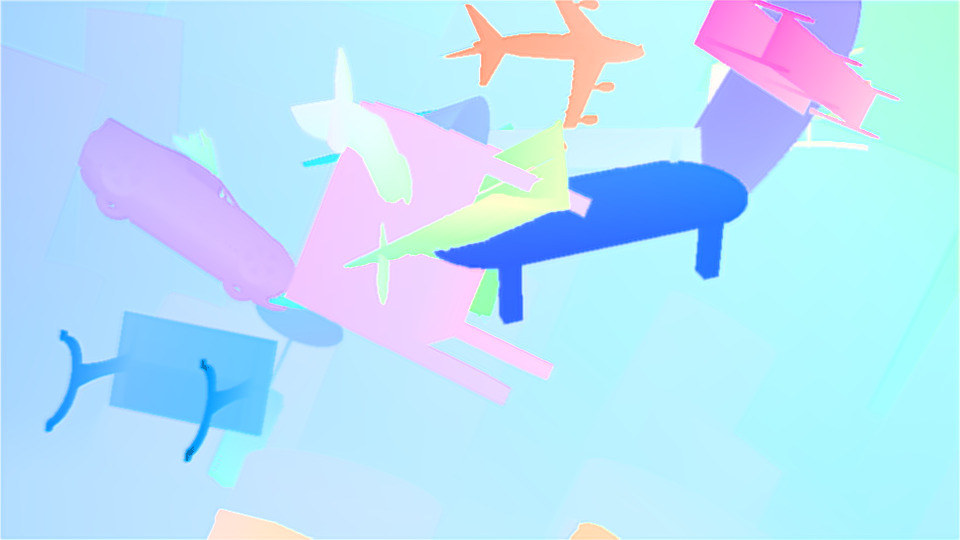}
		\caption{Inverted Optical Flow}
		\vspace{1.5mm}
	\end{subfigure}
	\begin{subfigure}[c]{0.45\linewidth}
		\includegraphics[width=\linewidth]{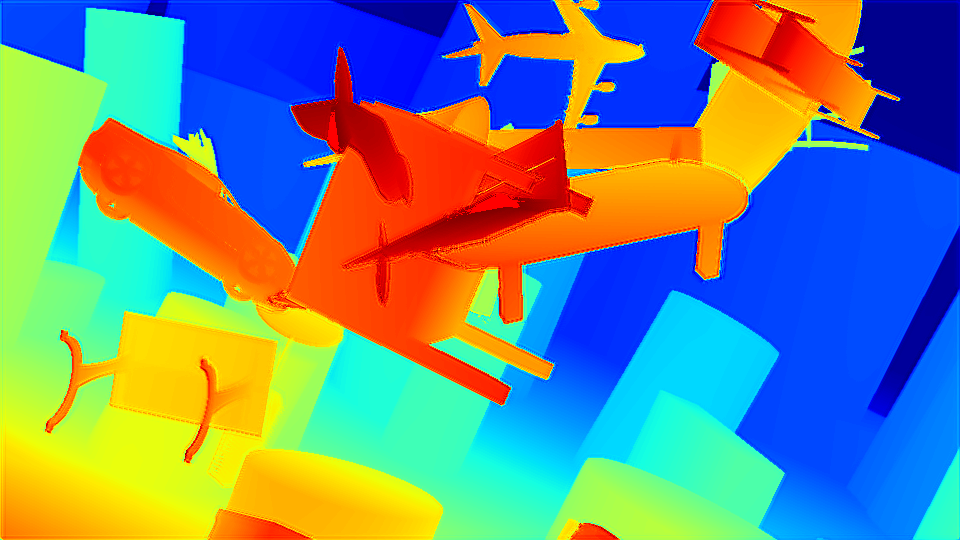}
		\caption{Inverted Disparity at $t+1$}
		\vspace{1.5mm}
	\end{subfigure}\\%
	\begin{subfigure}[c]{0.45\linewidth}
		\includegraphics[width=\linewidth]{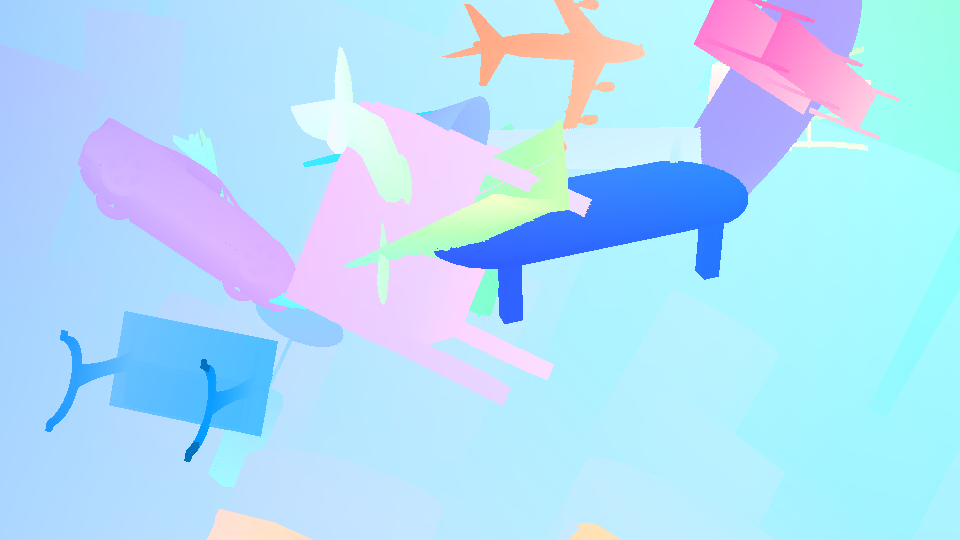}
		\caption{Forward Optical Flow}
	\end{subfigure}
	\begin{subfigure}[c]{0.45\linewidth}
		\includegraphics[width=\linewidth]{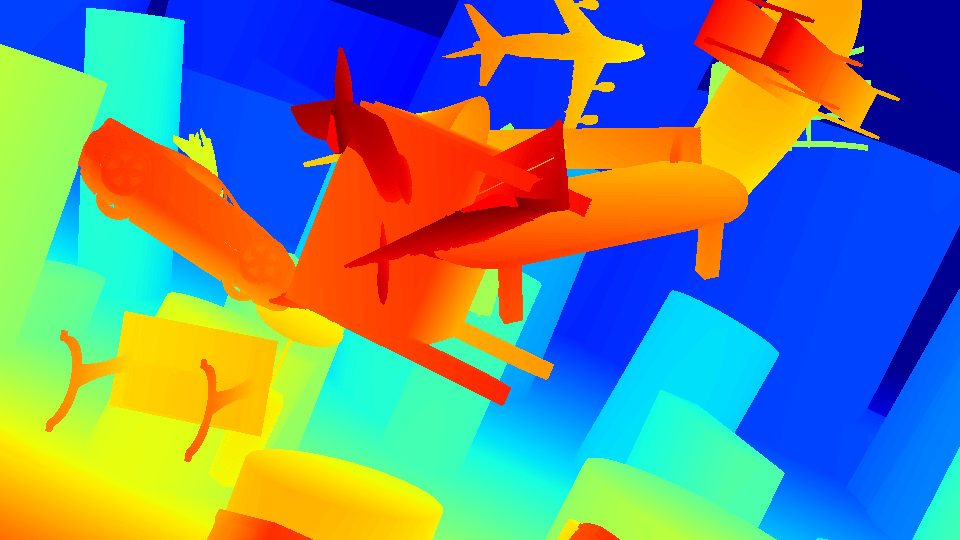}
		\caption{Disparity at $t+1$}
	\end{subfigure}
	\caption{An example of the learned inversion of motion on data of FlyingThings3D \cite{mayer2016large}. The left and right columns show the optical flow and disparity at $t+1$ components of the scene flow. The first and last rows give the ground truth in backward and forward direction respectively. The center row presents the results of our generic motion inverter.}
	\label{fig:inverter}
\end{figure}

Instead of a constant motion assumption, which is often applied in previous work, we create and train a compact neural network that utilizes a learned motion model to temporally invert scene flow.
Our architecture is inspired by the inversion module of \cite{maurer2018proflow} but we make it deeper since for our framework we want to learn a generic model that can invert motion for arbitrary sequences without the need of re-training on every frame.
In detail, the inversion sub-network consists of 4 convolutional layers with kernel size $3 \times 3$ and a fifth one with a $7 \times 7$ kernel and output feature dimensions of $16, 16, 16, 16, 4$ respectively.
The last layer is activated linearly.
Similarly to \cite{maurer2018proflow}, we equip our inverter with a mechanism for spatial variance by concatenating the input scene flow with normalized ($[-1, 1]$) spatial image coordinates of x- and y-direction.
This way and together with the depth information from the backward scene flow, the inversion network is able to operate fully in (hypothetical) 3D space.
For a qualitative impression of our inverter, \cref{fig:inverter} visualizes the results for a validation sample.

\subsection{Deep Forward-Backward Fusion} \label{sec:method:merger}
After the prediction of scene flow in the forward and backward direction (using the same reference frame) and inverting the backward estimate, we can merge the forward and inverted backward prediction.
The refined results can potentially overcome errors in difficult regions of occlusion or out-of-view motion, because occlusions occur rarely across multiple views \cite{schuster2020sffpp}.
Our fusion strategy follows a weighted average approach, where a fusion module predicts pixel-wise weights (that sum up to one) for the combination of the original forward estimate and the inverted backward scene flow.
Interestingly, these weights correspond to (soft) occlusion masks, revealing the main reason why the inverted backward motion should be preferred over a forward dual-frame estimate (cf. \cref{fig:teaser,fig:framework}).
While the direct prediction of a refined (or residual) scene flow during fusion is also possible, this would neither model the underlying issue nor produce occlusion masks.

For our fusion module, we adopt the architecture of the context network of PWC-Net \cite{sun2018pwc} and PWOC-3D \cite{saxena2019pwoc}.
It consists of seven convolutional layers with a kernel size of $3 \times 3$, output depth of $32, 64, 128, 128, 64, 32, 2$, and dilation rates of $1, 2, 4, 8, 16, 1, 1$ respectively.
The last layer predicts pseudo probabilities in a one-hot encoding for the forward and inverted backward scene flow which are used for weighted averaging after a softmax activation.
As input for this module, we concatenate the forward and inverted backward estimate.

Described above is a simple baseline for temporal fusion of scene flow (\textit{basic}).
Within the experiments in \cref{sec:experiments:ablation} we will compare different variants of our fusion module.
Though the network can detect occlusion based on the depth (disparity) and motion of neighboring pixels, it can not estimate out-of-view motion without knowing where the field of view ends. 
This information could be guessed from the padding during convolution, however for more explicit modeling we again feed additional spatial information to the module, similar as with the inverter. We denote this variant as \textit{spatial}.
Another variant is again motivated by the issue of occlusion. 
Since in multiple views different parts of a reference image are occluded, we argue that the predicted occlusion masks (fusion weights) should differ for the different components of the scene flow, \eg between left and right view of a stereo camera, there are no occlusions due to motion.
Therefore this variant is predicting a separate occlusion map for each channel of our scene flow representation (in image space) and is depicted as \textit{4ch} since it predicts fusion weights for four scene flow channels (two for optical flow and two for initial and future disparities).
Lastly, we combine both strategies and name the combination \textit{spatial-4ch}.
In \cref{fig:framework,fig:teaser}, the occlusion maps (fusion weights) for the \textit{basic} variant are shown for the sake of clarity and space.

\section{Experiments} \label{sec:experiments}
Our experiments and results are split into three sets with the following main intentions.
First of all, we validate that the overall framework improves over the initial dual-frame estimates of different auxiliary scene flow models.
Secondly, we compare our work to existing multi-frame scene flow algorithms using the official public KITTI benchmark \cite{geiger2012kitti,menze2015object}.
Lastly, our goal is to validate each step of our framework separately by means of an extensive ablation study.

As metric, the common KITTI outlier rate is used which classifies per-pixel estimates as outliers if they deviate more than 3 px and 5~\% from the ground truth.
This metric is computed for the different components of our scene flow, \ie initial disparity (\textit{D1}), next disparity (\textit{D2}), optical flow (\textit{OF}), or for the entire scene flow (\textit{SF}) if either of the three previous components is classified as an outlier.
All outlier rates are averaged over all valid ground truth pixels of the respective data split.

\begin{table*}[t]
	\centering
	\caption{Comparison of our multi-frame fusion approach to the dual-frame results of the underlying auxiliary scene flow estimator for the entire image (\textit{all}) and occluded areas only ($\mathit{occ} \in \mathit{all} \setminus \mathit{noc}$) on our KITTI validation split. The last column gives the maximum relative improvement of DTF over the respective dual-frame baseline.}
	\label{tab:dual}
	%\resizebox{1\linewidth}{!}{
	\begin{tabular}{cc||cccc|cccc|c}
		\multirow{2}{*}{\begin{tabular}{c}Scene Flow\\Estimator\end{tabular}} & \multirow{2}{*}{Setup} & \multicolumn{4}{c|}{all} & \multicolumn{4}{c|}{occ} & \multirow{2}{*}{\begin{tabular}{c}max. rel.\\Improv.\end{tabular}}\\
		 & & D1 & D2 & OF & SF & D1 & D2 & OF & SF\Bstrut\\
		\hline
		\hline
\multirow{2}{*}{SENSE \cite{jiang2019sense}} & Dual & 0.97 & 2.22 & 3.00 & 4.04 & 2.08 & 8.23 & 7.19 & 11.84&\Tstrut\\
 & \textbf{Ours} & \cellcolor{limegreen!0}0.97 & \cellcolor{limegreen!50}1.66 & \cellcolor{red!1}3.01 & \cellcolor{limegreen!42}3.52 & \cellcolor{limegreen!4}2.05 & \cellcolor{limegreen!50}4.81 & \cellcolor{red!0}7.21 & \cellcolor{limegreen!50}8.57 & 41.6 \%\Bstrut\\
\hline
\multirow{2}{*}{OE \cite{yang2020upgrading}} & Dual & 1.11 & 2.58 & 5.56 & 6.61 & 2.53 & 7.34 & 15.06 & 17.73&\Tstrut\\
 & \textbf{Ours} & \cellcolor{red!3}1.12 & \cellcolor{limegreen!15}2.46 & \cellcolor{limegreen!5}5.46 & \cellcolor{limegreen!11}6.39 & \cellcolor{red!1}2.54 & \cellcolor{limegreen!16}6.97 & \cellcolor{limegreen!10}14.57 & \cellcolor{limegreen!16}16.86 & 5.0 \%\Bstrut\\
\hline
\multirow{2}{*}{DWARF \cite{aleotti2020dwarf}} & Dual & 2.35 & 3.49 & 7.07 & 8.16 & 3.94 & 7.59 & 17.70 & 19.63&\Tstrut\\
 & \textbf{Ours} & \cellcolor{limegreen!50}1.17 & \cellcolor{limegreen!50}2.63 & \cellcolor{limegreen!50}5.64 & \cellcolor{limegreen!50}6.75 & \cellcolor{limegreen!50}2.82 & \cellcolor{limegreen!2}7.54 & \cellcolor{limegreen!50}14.90 & \cellcolor{limegreen!30}17.82 & 50.2 \%\Bstrut\\
\hline
\multirow{2}{*}{PWOC-3D \cite{saxena2019pwoc}} & Dual & 4.65 & 6.72 & 11.50 & 13.64 & 8.02 & 15.20 & 29.17 & 32.15&\Tstrut\\
 & \textbf{Ours} & \cellcolor{limegreen!50}3.34 & \cellcolor{limegreen!50}4.85 & \cellcolor{limegreen!50}8.22 & \cellcolor{limegreen!50}9.70 & \cellcolor{limegreen!50}5.63 & \cellcolor{limegreen!50}10.10 & \cellcolor{limegreen!50}18.68 & \cellcolor{limegreen!50}21.24 & 36.0 \%\Bstrut\\
\hline
\multirow{2}{*}{SFF \cite{schuster2018sceneflowfields}} & Dual & 6.61 & 10.28 & 12.39 & 15.76 & 9.94 & 19.57 & 26.08 & 30.74&\Tstrut\\
 & \textbf{Ours} & \cellcolor{limegreen!28}6.04 & \cellcolor{limegreen!40}9.03 & \cellcolor{limegreen!25}11.43 & \cellcolor{limegreen!30}14.30 & \cellcolor{limegreen!39}8.77 & \cellcolor{limegreen!50}15.91 & \cellcolor{limegreen!41}22.85 & \cellcolor{limegreen!48}26.25 & 18.7 \%\Bstrut\\

	\end{tabular}
	%}
\end{table*}

\begin{figure*}[t]
	\centering
	\begin{subfigure}[c]{0.3\linewidth}
		\includegraphics[width=\linewidth]{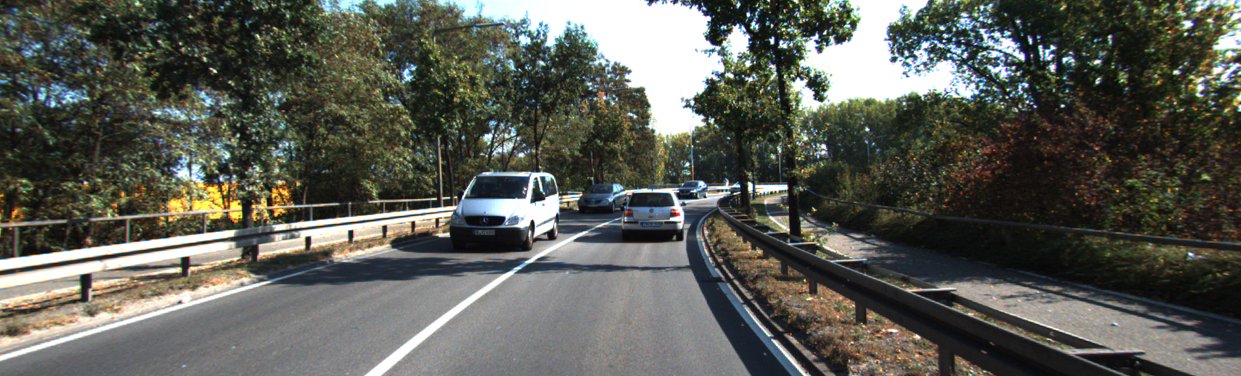}
		\caption{Reference Image}
		\vspace{1.5mm}
	\end{subfigure}
	\begin{subfigure}[c]{0.3\linewidth}
		\includegraphics[width=\linewidth]{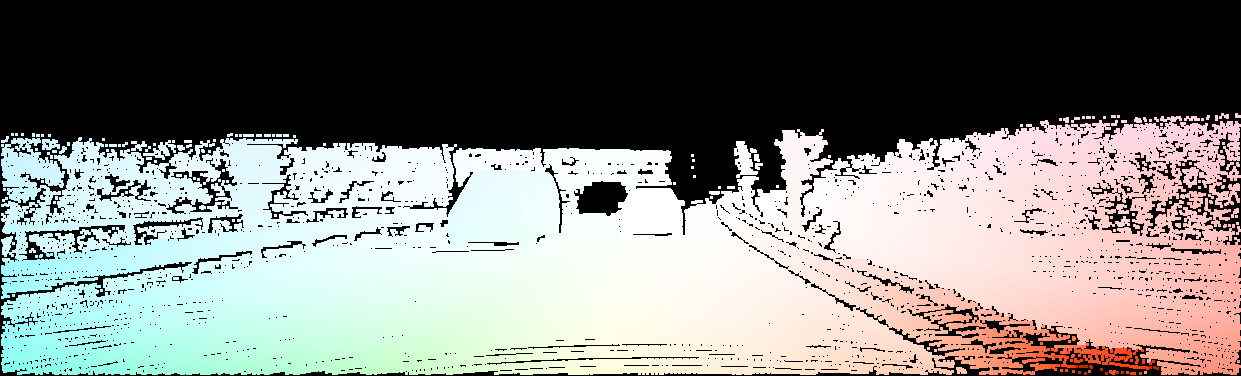}
		\caption{\textbf{Forward} Ground Truth (enhanced)}
		\vspace{1.5mm}
	\end{subfigure}\\%
	\begin{subfigure}[c]{0.3\linewidth}
		\includegraphics[width=\linewidth]{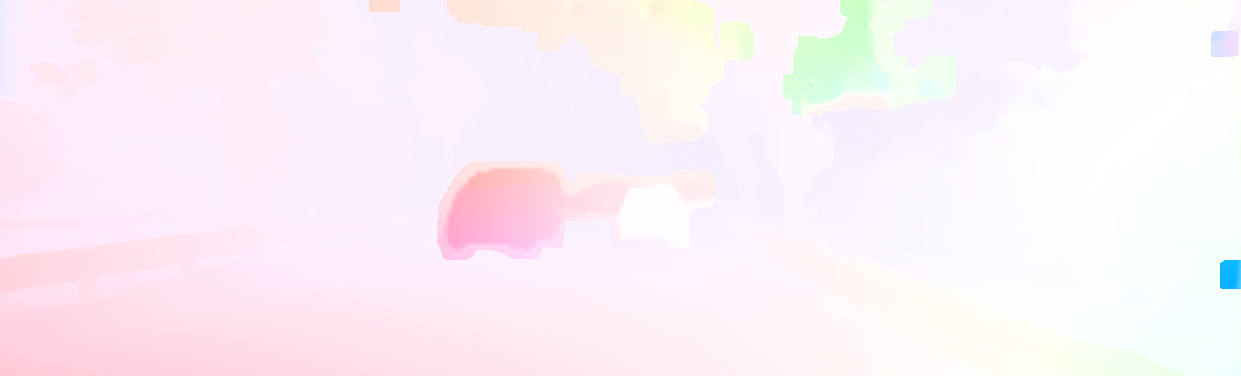}
		\caption{SENSE \cite{jiang2019sense}}
		\vspace{1.5mm}
	\end{subfigure}
	\begin{subfigure}[c]{0.3\linewidth}
		\includegraphics[width=\linewidth]{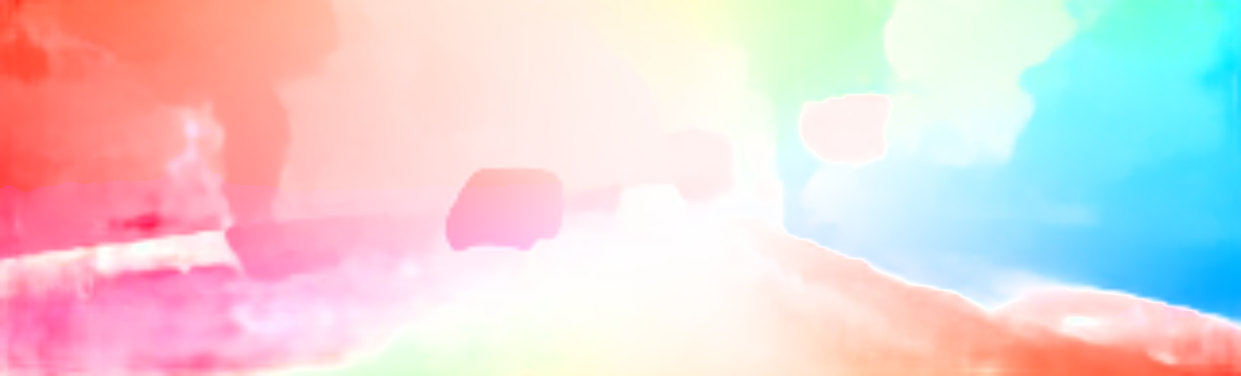}
		\caption{OpticalExpansion \cite{yang2020upgrading}}
		\vspace{1.5mm}
	\end{subfigure}
	\begin{subfigure}[c]{0.3\linewidth}
		\includegraphics[width=\linewidth]{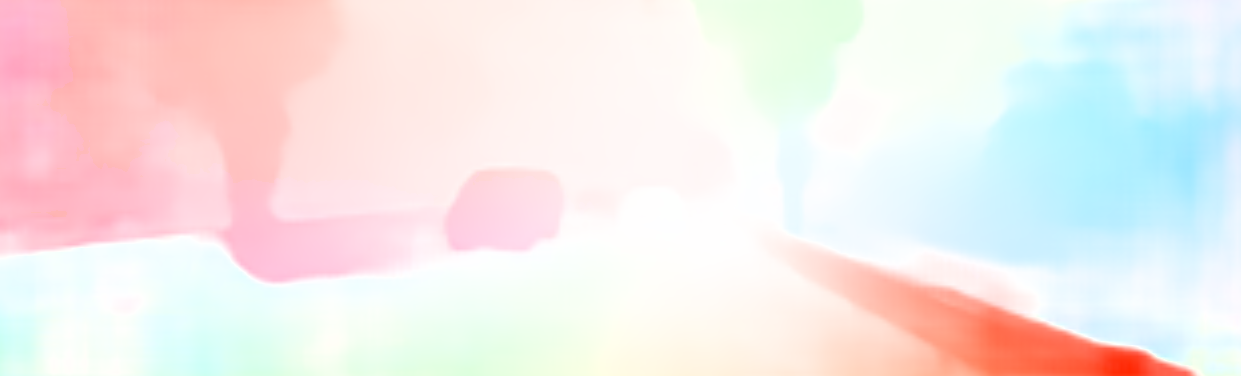}
		\caption{DWARF \cite{aleotti2020dwarf}}
		\vspace{1.5mm}
	\end{subfigure}\\%
	\begin{subfigure}[c]{0.3\linewidth}
		\includegraphics[width=\linewidth]{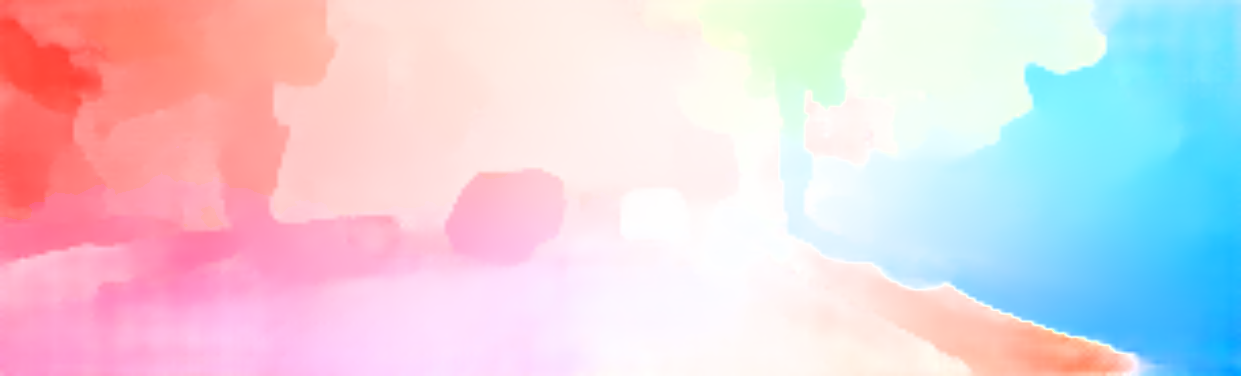}
		\caption{PWOC-3D \cite{saxena2019pwoc}, Original}
	\end{subfigure}
	\begin{subfigure}[c]{0.3\linewidth}
		\includegraphics[width=\linewidth]{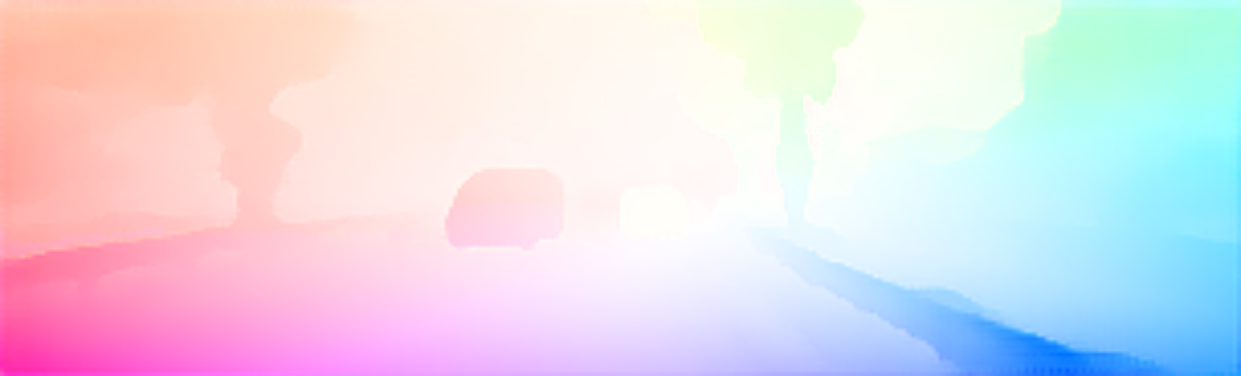}
		\caption{PWOC-3D \cite{saxena2019pwoc}, Re-trained}
		\label{fig:backward:retrained}
	\end{subfigure}
	\begin{subfigure}[c]{0.3\linewidth}
		\includegraphics[width=\linewidth]{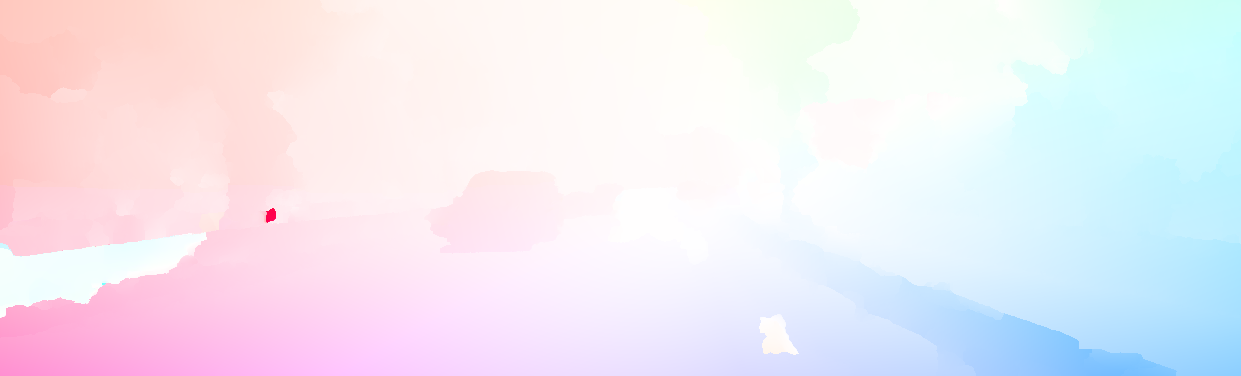}
		\caption{SFF \cite{schuster2018sceneflowfields}}
	\end{subfigure}	
	\caption{Visualization of the \textbf{backward} optical flow for different scene flow estimators. Most auxiliary estimators used in our experiments have difficulties with backward motion because they do not perform actual matching but rather rely on the image information of the reference frame alone, especially for street surfaces. Significant improvements are noticeable once the backward branch gets trained end-to-end in our framework (\subref{fig:backward:retrained}), even though backward ground truth is not available.}
	\label{fig:backward}
\end{figure*}

\subsection{Data Sets and Training} \label{sec:experiments:details}

\paragraph*{Data Sets.}
For most of our experiments, the well-known KITTI data set is used \cite{geiger2012kitti,menze2015object}.
However, it is limited in size and thus inappropriate for the training of deep neural networks.
Despite some success on unsupervised scene flow estimation \cite{hur2020selfmono} or knowledge distillation from teacher networks \cite{aleotti2020dwarf,jiang2019sense}, transfer learning by pre-training and fine-tuning is the most common strategy to overcome this issue \cite{mayer2018what,saxena2019pwoc,sun2018models,sun2018pwc}.
The one large-scale data set which provides labeled data for scene flow is FlyingThings3D (FT3D) \cite{mayer2016large}.
In this work, it is also used for pre-training of some parts of the pipeline.

For validation, we split 20 sequences from the KITTI \textit{training} subset as in \cite{saxena2019pwoc} and the last 50 sequences from each subset \textit{A}, \textit{B}, and \textit{C} of the FlyingThings3D \textit{train} set.

\paragraph*{Training and Implementation Details.}
Where required, the auxiliary scene flow estimators are initialized with the published pre-trained weights.
We use the rich ground truth of FlyingThings3D \cite{mayer2016large} to separately pre-train the inverter on forward and backward ground truth motion with an L2-loss for 40 epochs with a batch size of 4 and an initial learning rate of $1\times10^{-4}$ that we decrease to $5\times10^{-5}$ and $1\times10^{-5}$ after 20 and 30 epochs respectively.
The rest of our pipeline is initialized from scratch.

Afterwards, we fine-tune our fusion pipeline on KITTI \cite{menze2015object} for 100 epochs. The learning rate for fine-tuning starts at $5\times10^{-5}$ and is again reduced after 75 epochs to $1\times10^{-5}$.
Due to memory limitations, we use a batch size of 1 whenever the entire pipeline is used for training.

Unless mentioned otherwise, Leaky-ReLU \cite{maas2013leakyrelu} with a leak factor of $0.1$ is used after each convolution.
For all training stages, we use the ADAM optimizer \cite{kingma2015adam} with its default parameters.

Our robust loss function for the 4-dimensional scene flow in image space is similar to the one in \cite{saxena2019pwoc,sun2018pwc} and defined by
\begin{equation}
	\mathcal{L} = \frac{1}{N_{gt}} \cdot \sum_{\mathbf{x} \in gt} \left( \vert s(\mathbf{x}) - \hat{s}(\mathbf{x}) \vert_1 + \epsilon \right)^{0.4}.
\end{equation}
Here $s$ and $\hat{s}$ are the estimated and ground truth scene flow fields, $\vert \cdot \vert_1$ is the L$_1$-norm, $\epsilon=0.01$ is a small constant for numerical stability, and the power of $0.4$ gives less weight to strong outliers.

For the entire pipeline, we impose this loss on the forward estimate, the inverted backward scene flow, and the final fusion:
\begin{equation}
	\mathcal{L}_{total} = \mathcal{L}_{fw} + \mathcal{L}_{inv} + \mathcal{L}_{fused}
\end{equation}
This multi-stage loss avoids that during training the fusion flips to one side and does not recover because the other side would not receive any updates anymore.

\begin{figure*}[t]
	\centering
	\includegraphics[width=1\linewidth]{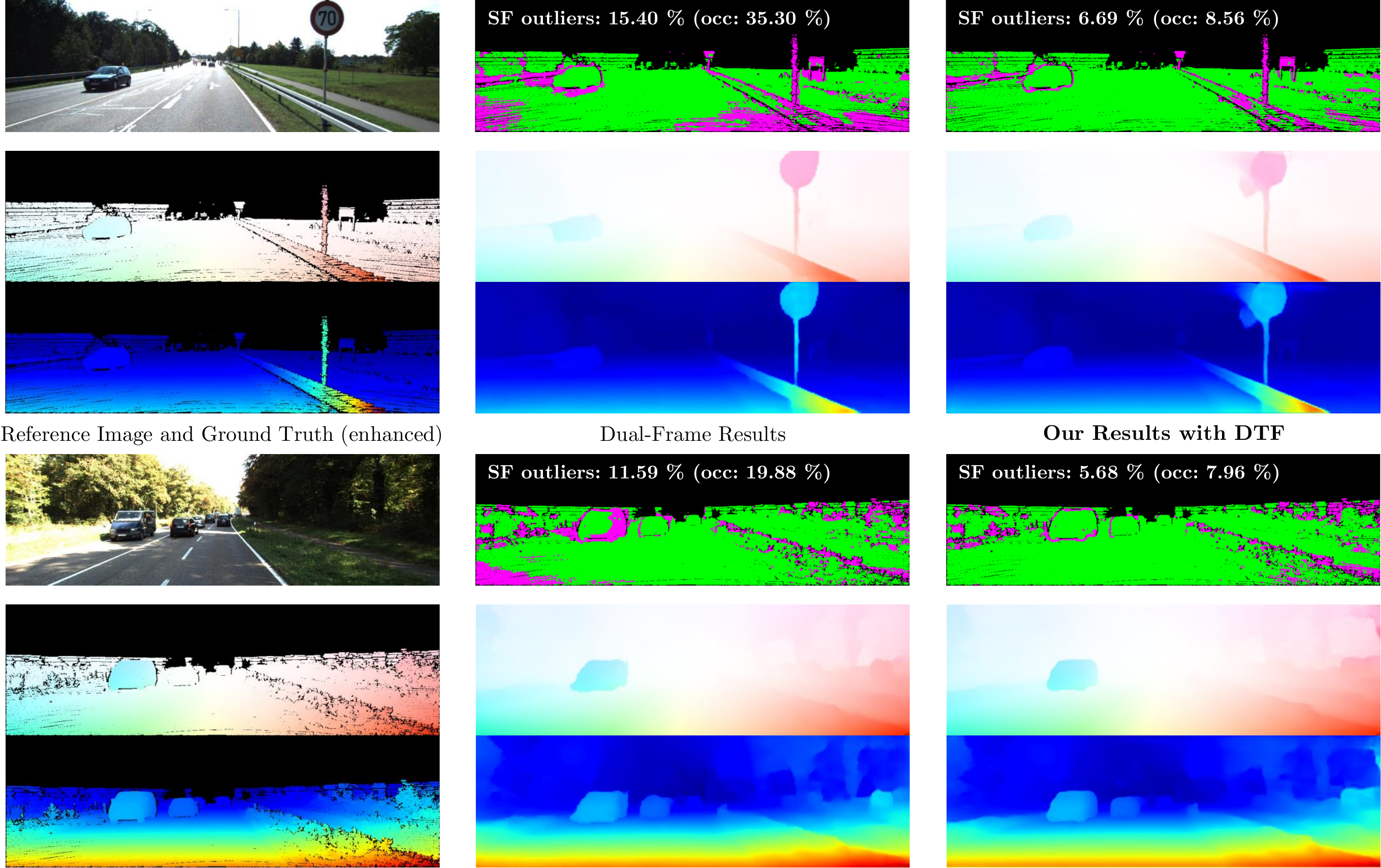}
	\caption{Visual comparison of our deep multi-frame fusion framework to the auxiliary dual-frame model PWOC-3D \cite{saxena2019pwoc}. Scene flow results are shown by optical flow and disparity at $t+1$. The error maps indicate scene flow outliers in magenta and inliers in green. Notice the improvements in occluded areas (\eg in front and around of vehicles) or the out-of-view occlusions due to ego-motion (\eg the close-by part of the guardrail in the first example and the lower image corners).}
	\label{fig:comparison}
\end{figure*}

\subsection{Comparison to the Auxiliary Estimators}  \label{sec:experiments:dual}
In \cref{tab:dual} we validate that our deep temporal fusion framework surpasses a diverse set of underlying dual-frame estimators in terms of scene flow outliers.
Especially in the difficult areas of occlusion, our approach achieves significantly better results, reducing the scene flow outlier rate by up to {\raise.17ex\hbox{$\scriptstyle\sim$}} 30~\%.
The fusion improves the scene flow estimates for non-occluded areas also, resulting in an overall improvement over \textit{all} image areas.
For OpticalExpansion (OE) \cite{yang2020upgrading}, the relative improvement is less compared to other auxiliary estimators.
This has two reasons. First of all, some scene flow algorithms are heavily biased towards forward motions (cf. \cref{fig:backward}) and therefore provide much less reliable information for fusion in the backward branch. Secondly, the estimate of motion-in-depth from OE is depending a lot on the optical flow estimate, which amplifies the previous limitation and expands it over the complete scene flow estimation in backward direction.
The first reason additionally motivates an end-to-end training of the fusion framework together with the auxiliary estimator.
This is performed for PWOC-3D \cite{saxena2019pwoc} because it is most easy to train.
The other auxiliary estimators are used as off-the-shelf replacements with the officially provided pre-trained weights.
Our framework is even able to improve non-learning-based results from SFF \cite{schuster2018sceneflowfields}, with a noticeable margin of more than 10~\% in occluded areas.
Here, we account the smaller relative improvements to the ego-motion model that is applied in SFF which is able to estimate out-of-view motions in forward direction for the background more reliably.
A visual comparison between PWOC-3D and the multi-frame extension by our framework is conducted in \cref{fig:comparison}.

\begin{table*}
	\centering
	\caption{Results of the KITTI scene flow benchmark for all multi-frame approaches. We also provide results for the auxiliary scene flow methods used in our pipeline and conceptual dual-frame counterparts for other multi-frame methods, where existent. Scene flow outlier rates (\textit{SF}) are presented for foreground (\textit{fg}), background (\textit{bg}), and all regions, as well as for non-occluded areas (\textit{noc}), occluded areas only (\textit{occ}, details in the text), and the union (\textit{all}).}
	\label{tab:kitti}
	%\resizebox{1\linewidth}{!}{
	\begin{tabular}{c|c||ccc|ccc|ccc|c}
		 & \multirow{3}{*}{Method} & \multicolumn{9}{c|}{SF Outliers [\%]} & \multirow{3}{*}{\begin{tabular}{c}Run\\Time\\{[}s{]}\end{tabular}}\\
		 & & \multicolumn{3}{c|}{occ} & \multicolumn{3}{c|}{noc} & \multicolumn{3}{c|}{all} &\\
		 & & bg & fg & all & bg & fg & all & bg & fg & all &\Bstrut\\
		\hline
		\hline
		\multirow{6}{*}{\rotatebox[origin=c]{90}{multi-frame}} & PRSM \cite{vogel2015PRSM} & \textbf{12.36} & 37.65 & \textbf{15.74} & 5.54 & 17.65 & 7.71 & \textbf{6.61} & 20.79 & \textbf{8.97} & 300\Tstrut\\
		 & DTF+SENSE (\textbf{Ours}) & 16.37 & \textbf{37.49} & 19.65 & 6.69 & \textbf{9.72} & \textbf{7.23} & 8.21 & \textbf{14.08} & 9.18 & 0.76\\
		 & OSF+TC \cite{neoral2017object} & 15.46 & 43.98 & 19.49 & \textbf{5.52} & 15.57 & 7.32 & 7.08 & 20.03 & 9.23 & 3000\\
		 & SFF++ \cite{schuster2020sffpp} & 26.40 & 48.36 & 30.91 & 9.84 & 21.04 & 11.55 & 12.44 & 25.33 & 14.59 & 78\\
		 & DTF+PWOC (\textbf{Ours}) & 31.91 & 51.14 & 34.29 & 8.79 & 21.01 & 10.98 & 12.42 & 25.74 & 14.64 & \textbf{0.38}\\
		 & FSF+MS \cite{taniai2017fsf} & 21.59 & 65.48 & 27.63 & 9.23 & 28.03 & 12.60 & 11.17 & 33.91 & 14.96 & 2.7\Bstrut\\
		\hline
		\multirow{5}{*}{\rotatebox[origin=c]{90}{dual-frame}} & SENSE \cite{jiang2019sense} & 17.22 & 44.86 & 21.63 & 6.71 & 10.02 & 7.30 & 8.36 & 15.49 & 9.55 & 0.32\Tstrut\\
		 & OSF \cite{menze2015object} & 15.01 & 47.98 & 19.41 & 5.52 & 22.31 & 8.52 & 7.01 & 26.34 & 10.23 & 3000\\
		 & PWOC-3D \cite{saxena2019pwoc} & 41.20 & 47.52 & 41.62 & 9.29 & 18.03 & 10.86 & 14.30 & 22.66 & 15.69 & 0.13\\
		 & SFF \cite{schuster2018sceneflowfields} & 25.58 & 63.26 & 30.76 & 10.04 & 26.51 & 12.99 & 12.48 & 32.28 & 15.78 & 65\\
		 & PRSF \cite{vogel2013PRSF} & 41.09 & 58.82 & 42.80 & 8.35 & 26.08 & 11.53 & 13.49 & 31.22 & 16.44 & 150\\
	\end{tabular}
	%}
\end{table*}

\begin{table*}
	\centering
	\caption{Evaluation of intermediate results in our pipeline on our KITTI validation split. For this experiment, PWOC-3D \cite{saxena2019pwoc} is the auxiliary estimator and is trained end-to-end. The inversion module is separately evaluated on FlyingThings3D.}
	\label{tab:ablation}
	%\resizebox{1\linewidth}{!}{
	\begin{tabular}{l||cccc|cccc}
	 	\multirow{2}{*}{Output} & \multicolumn{4}{c|}{all} & \multicolumn{4}{c}{occ}\\
		 & D1 & D2 & OF & SF & D1 & D2 & OF & SF\Bstrut\\
		\hline
		\hline
		forward (fw) & 3.47 & 5.83 & 8.95 & 10.76 & 5.89 & 14.39 & 23.17 & 26.93\Tstrut\\
		inverted backward (bw-inv) & 4.15 & 6.00 & 20.34 & 22.14 & 6.64 & 9.92 & 31.74 & 33.81\Bstrut\\
		\hline
		constant linear inversion (FT3D) & -- & \textbf{1.27} & 47.16 & 47.18 & -- & -- & -- & --\Tstrut\\
		our inverter (FT3D) & 2.19 & 3.25 & \textbf{41.98} & \textbf{42.34} & -- & -- & -- & --\Bstrut\\
		\hline
		fw + bw-inv + oracle & 2.63 & 3.91 & 6.25 & 7.51 & 4.53 & 8.40 & 16.39 & 18.43\Tstrut\Bstrut\\
		\hline
		fw + bw-inv + fusion-basic & \textbf{3.22} & 4.90 & 9.01 & 10.48 & 4.88 & 10.23 & 19.27 & 21.66\Tstrut\\
		fw + bw-inv + fusion-spatial & 3.48 & 5.51 & 8.85 & 10.55 & 6.13 & 13.66 & 22.23 & 25.40 \\
		fw + bw-inv + fusion-4ch & 3.34 & 4.85 & \textbf{8.22} & \textbf{9.70} & 5.63 & 10.10 & 18.68 & 21.24 \\
		fw + bw-inv + fusion-spatial-4ch & 3.43 & \textbf{4.84} & 8.67 & 10.19 & \textbf{5.45} & \textbf{9.25} & \textbf{18.46} & \textbf{20.82} \\
	\end{tabular}
	%}
\end{table*}

\subsection{Comparison to State-of-the-Art} \label{sec:experiments:sota}
To check the generalization of our model on more unseen data, we submit results obtained with our deep multi-frame model to the KITTI online benchmark.
The results for all multi-frame methods and related dual-frame baselines are presented in \cref{tab:kitti}.
Due to the limited number of training samples on KITTI, some over-fitting can be observed when comparing the numbers to the results on our validation split.
However, improvements over the underlying dual-frame models (SENSE and PWOC-3D) are still evident, again with margins of {\raise.17ex\hbox{$\scriptstyle\sim$}} 15 - 20~\% in occluded areas.
Since KITTI evaluates the submitted results only for non-occluded (\textit{noc}) and all valid pixels, the results for occluded areas (\textit{occ}) are reconstructed from the available data.
To this end, we compute the ratio of non-occluded image areas on the KITTI \textit{training} set (84.3 \%), and use this distribution to estimate results for only occluded areas for the KITTI \textit{testing} set based on the benchmark results for non-occluded (\textit{noc}) and \textit{all} areas according to the following formula:
\begin{equation}
	{occ}_r = \frac{{all}_r - {noc}_r \cdot 0.843}{0.157}
\end{equation}
for the regions $r \in \lbrace bg, fg, all \rbrace$.
This strategy reveals that even for the top performing multi-frame methods, moving vehicles which leave the field of view are the most challenging areas.
In these regions (\textit{occ-fg}), our fusion approach achieves top performance.
It furthermore performs significantly better in foreground regions than the other multi-frame methods.
Lastly, we highlight that since ours is the first deep method for multi-view scene flow estimation, our run time is close-to real time and thus 2 to 5 orders of magnitude faster than that of most other multi-view methods.
The inversion and fusion without auxiliary scene flow estimation takes 0.12 seconds.
We use a Nvidia RTX 2080 Ti for inference. 

\subsection{Ablation Study} \label{sec:experiments:ablation}
For completeness, each part of our framework is evaluated separately in \cref{tab:ablation}.
The first two rows show the results for the forward prediction and the inverted backward scene flow after end-to-end training.
We can see that within our multi-view training, the plain forward prediction is already improved over the dual-frame baseline (cf. \cref{tab:dual}).
Further, the results of the backward branch after inversion indicate that the motion inversion of optical flow is a bottleneck.
Yet, for occluded areas the inversion outperforms the forward prediction already in terms of change of disparity, validating its importance.
Both of these observations are confirmed by an evaluation of the inverter only on data of FlyingThings3D \cite{mayer2016large} as shown in the fourth row of \cref{tab:ablation} (cf. \cref{fig:inverter}) compared to a na\"ive constant linear motion assumption in 2D.
This is, optical flow and change of disparity are multiplied by $-1$.
Our learned motion model outperforms the constant motion model in terms of optical flow.
Though, one might doubt whether the quality of the inversion is good enough to improve the forward prediction.
Therefore, we compute an \textit{oracle} fusion using the ground truth to select the better estimate from the forward and inverted backward branch.
This experiment produces a theoretical bound for our fusion module and makes apparent that the inverted backward scene flow contains a lot of valuable information.
Within the last four rows of \cref{tab:ablation} we compare the different variants of our fusion module as described in \cref{sec:method:merger}.
The results in occluded areas reveal that all variants including the \textit{basic} one effectively tackle the problem of occlusion.
Among all, the \textit{spatial} version performs the worst unless combined with the \textit{4ch} variant.
However, we could observe stronger over-fitting for this model with most representation power (and highest number of parameters).
As a result, over the entire image area, the fusion module using four weight channels performs the best.
Worth highlighting is that our fusion results in occluded areas reach the level of the oracle prediction almost.

\section{Conclusion} \label{sec:conclusion}
In this work we have presented a straight-forward integration of multiple frames to improve scene flow estimates for a wide range of dual-frame algorithms.
Significant improvements could be achieved by inverting the backward motion of the reference view and fusing it with an initial forward estimate.
Moreover, our fusion strategy of weighted averages yields additional estimates of occlusion maps without the need for bi-directional consistency checks.

The experiments reveal that the inversion of optical flow is a limiting factor of the proposed approach, thus for future work we plan to equip the motion inverter with more domain knowledge to overcome this limitation and further to apply end-to-end training with other more complicated auxiliary estimators.

\section*{Acknowledgement}
This work was partially funded by the BMW Group and partially by the Federal Ministry of Education and Research Germany under the project VIDETE (01IW18002).

{\small
\bibliographystyle{ieee_fullname}
\bibliography{bib}
}

\end{document}